\documentclass[12pt]{article}
\usepackage[margin=1.2in]{geometry}
\usepackage[T1]{fontenc}
\usepackage{amsmath,amssymb,booktabs,tabularx,array}
\usepackage{newtxtext,newtxmath}
\usepackage{microtype,graphicx,setspace,pifont,multirow}
\usepackage[backend=biber,style=ieee,citestyle=numeric-comp,sorting=none,
    natbib=true,doi=false,url=true,eprint=false,isbn=false,
    date=comp,urldate=long]{biblatex}
\usepackage{xpatch}
\xpatchbibdriver{online}
    {\printlist{organization}}
    {\printlist{organization}\newunit\newblock
     \iffieldundef{year}{\printtext{\bibstring{nodate}}}{\usebibmacro{date}}}
    {}{\PackageError{manuscript}{Could not add online publication dates}{Check the biblatex-ieee online driver.}}
\xpatchbibdriver{online}
    {\newunit\newblock\usebibmacro{url+urldate}}
    {\setunit{\adddot\space}\newblock\usebibmacro{url+urldate}}
    {}{\PackageError{manuscript}{Could not format online access information}{Check the biblatex-ieee online driver.}}
\DeclareFieldFormat[online]{version}{Version\addspace#1}
\usepackage{xcolor,caption,enumitem}
\usepackage{xurl}
\usepackage[hidelinks]{hyperref}
\setlist{itemsep=2pt,topsep=3pt}

\AtBeginBibliography{\singlespacing\interlinepenalty=10000\raggedright}

\newcommand{\D}{\mathsf{D}}
\newcommand{\Sh}{\mathsf{S}}
\newcommand{\Pass}{\mathsf{P}}
\newcommand{\Fail}{\mathsf{F}}

\newcommand{\yes}{\checkmark}
\newcommand{\xmark}{\ding{55}} 
\newcommand{\no}{\xmark}

\title{\resizebox{\textwidth}{!}{Game theory of academic effort and hiring in the age of AI}}
\author{Saptarshi Pal$^1$ and Peter S. Park$^2$}
\date{\large
    $^1$Department of Mathematics, Harvard University\\%
    $^2$Department of Human Evolutionary Biology, Harvard University\\[2ex]%
    \today
}

\begin{document}
\onehalfspacing
\maketitle

\section*{Abstract}

\noindent Many academic disciplines (in particular, mathematics and the theoretical sciences) face difficult questions about how to adapt their practices to the rise of highly capable artificial intelligence (AI).
Before AI, a well-written research paper with a strong theoretical result acted as a reliable signal that its authors understood their field in depth.
Now that AI is widely accessible, such signals have become cheaper to produce.
As a result, they no longer reliably sort academics by their level of understanding.
This has direct consequences for academic hiring. The age of AI raises a question for each side of the academic job market.
Should an early-career researcher invest in a deep understanding of their field, or instead use AI to build a longer resume with many papers?
Should a recruiter favor applicants with fewer papers who pass an interview aimed at probing understanding, or insist on both publication volume and a pass?
In this paper, we develop a game-theoretic model of this two-sided dilemma.
We classify all pure-strategy equilibria that can arise in this game and use learning simulations to identify the behaviors that populations of applicants and recruiters may learn to adopt.
Our analysis provides a framework for reasoning about how academic hiring can adapt to AI and for evaluating practices that preserve the value of deep understanding.

\clearpage

\section{Introduction}\label{sec:intro}

Academic publishing serves a dual purpose. First, it provides a lasting record of scientific discoveries and a means for scholars to share, evaluate, and build upon others' work. The second purpose is less explicit. Publications have long served as an indicator of an author's understanding of their field. A paper in a respected peer-reviewed journal was generally taken as evidence of expertise, often without closer examination. This particular aspect of publishing shapes academic job prospects for early-career academics. Graduate students and postdocs seek to publish in prestigious journals, motivated both by the scientific value of their work and by the recognition that publication brings. For an early-career scientist, publishing is thus both a means of advancing knowledge and a way to build the reputation needed to compete for academic positions. Academic recruiters, in turn, have long relied on strong publications as evidence of a candidate's depth of understanding.\\

\noindent With the advent of highly capable and easily accessible AI, however, producing high-quality research papers is no longer so rare. For example, AI models can now make progress on long-standing problems across several mathematical fields \citep{openainavierstokes, openai700, anthropicriemannzeta}. Their users can therefore produce impressive research without developing the depth of understanding that was once necessary to produce the same output \citep{tao2026}. This effect is most pronounced in mathematics and in the theoretical branches of science and engineering. Specifically in these domains, AI has weakened the once-tight link between producing strong results and understanding one's field. \\

\noindent In response to AI, academic communities are now reconsidering how they evaluate their researchers. A summit of mathematicians discussing the future of the math Ph.D. recommended placing less weight on dissertation texts, and more weight on in-person assessments and thesis defenses \citep{summit2026}. A working group in theoretical computer science proposed reviewing at most three works per applicant, and also recommended `chalk talks' to assess technical depth \citep{simons2026}. The free and open-access preprint server arXiv has introduced a monthly quota of two submissions per individual \citep{arxiv2026}. These measures echo calls to cap researchers' publication rates \citep{pasquato2026,craven2026,chu2026e}.\\

\noindent Other responses focus on research norms. Some mathematicians have pledged to work without AI \citep{beginproof2026}. Others oppose hiring colleagues who substitute AI output for their own work \citep{lazic2026,weinreich2026}. Joint statements and policies address attribution, disclosure, and AI's effects on mathematical practice \citep{leiden2026,roberts2026,iciam2026,siampolicy2026,mathathon2026}. Early-career mathematicians must navigate these changing expectations while building their careers \citep{taback2026}. At the same time, departments and other recruiters of researchers must decide how to evaluate applicants whose preparation responds to those very decisions.
Each side's best choice therefore depends on that of the other, and neither can be settled in isolation.
Game theory is designed to analyze exactly this kind of strategic interdependence.\\

\noindent In this paper, we study a game-theoretic model of the incentives on both sides of the academic job market in the age of AI. Applicants differ in potential and choose one of two trajectories: deep study or shallow study with AI. Deep study builds theoretical understanding but costs more to the applicant than shallow study with AI. Furthermore, deep study may reduce paper output for applicants who are not of high potential. Shallow study with AI is cheaper and produces high output without building understanding. How much more deep study costs also depends on the applicant's potential.
Recruiters observe only an applicant's paper output and the result of a possibly imperfect test of understanding, such as an interview, which the applicant either passes or fails (the likelihood to pass is much larger if an applicant \textit{understands}).
Because an applicant's potential and their study choice remain hidden from recruiters, recruiters must infer understanding from two signals: paper output and the applicant's performance in the test.
The recruiters' hiring decisions depend on how much they value output and understanding, and they may reward either one or both.
We ask how applicants of different potentials should choose their study trajectories, and how recruiters should decide when presented with a publication list and a test result.\\

\noindent We characterize all pure-strategy perfect Bayesian equilibria of this game. These equilibria identify patterns of study and hiring that neither side has an incentive to unilaterally change. We then examine which behaviors emerge under a learning process, such as payoff-based imitation and rare exploration. This evolutionary analysis helps distinguish possible equilibria from outcomes that tend to emerge through learning. We also relate proposed institutional reforms \cite{summit2026,simons2026,arxiv2026} to changes in the model's parameters. This yields conditional predictions about how those reforms could affect study and hiring.\\

\noindent Section~\ref{sec:game} defines the game, and Section~\ref{sec:equilibrium} characterizes its equilibria. Section~\ref{sec:learning} examines evolutionary selection among outcomes. Section~\ref{sec:discussion} discusses institutional implications, related work, and limitations. The Appendix (Section~\ref{sec:methods}) gives the evolutionary methods.

\section{Game-theoretic model of academic effort and hiring}

\label{sec:game}

\textbf{Game description.} We study an effort-and-hiring signaling game between applicants and recruiters. A candidate (say, a graduate student) shows up to a job interview with a resume that highlights the academic papers they have under their belt. We classify profiles with a binary variable $y$, with $y = 0$ denoting few papers and $y=1$ denoting many papers. We assume it is impossible to fake a profile. That is, a hiring committee can reliably check if a candidate is lying on their resume. \\

\begin{figure}[t!]
    \centering
    \includegraphics[width=\linewidth]{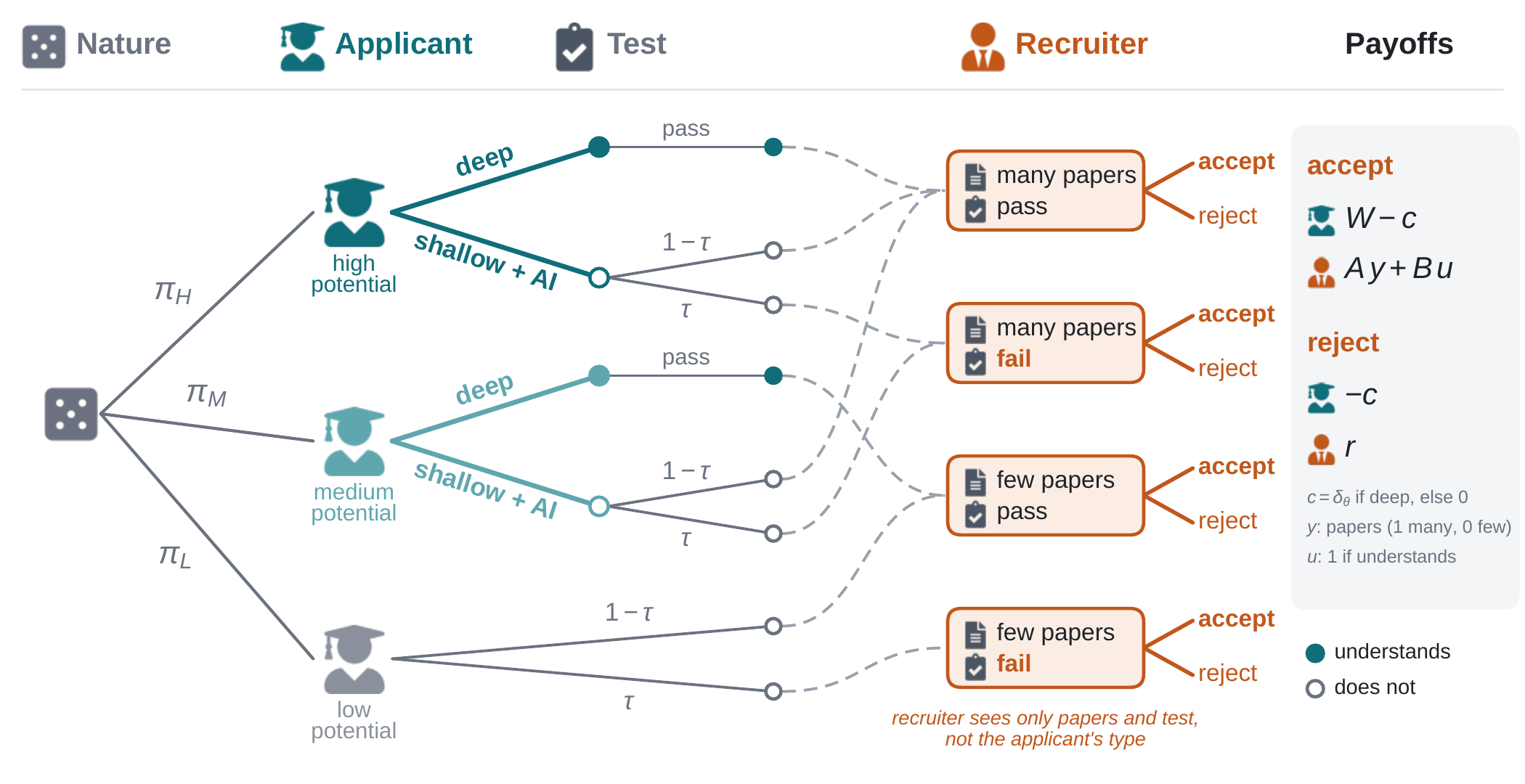}
    \caption{\onehalfspacing \textbf{The effort-and-hiring game.} In Section \ref{sec:game}, we describe the game in full detail. This figure illustrates the textual description. First, nature produces an applicant that is either high-potential, medium-potential, or low-potential, with probabilities $\pi_H, \pi_M$, and $\pi_L$, respectively (each $>0$). The high and medium-potential applicants then choose between two strategies: deep and shallow + AI. Choosing \textit{Deep} yields a deep understanding of the field ($u=1$), and choosing \textit{shallow + AI} produces a shallow understanding of the field ($u=0$). In addition to determining the understanding of an applicant, a strategic trajectory (deep or shallow) also affects the volume of academic papers an applicant produces pre-interview. In the baseline model, the high-potential individual produces many papers regardless of their strategy ($y=1$), but the medium-potential individual only produces many papers if they choose shallow + AI. By going deep, the medium-potential individual earns few or no papers ($y=0$). The low-potential type is non-strategic: they have no understanding and they produce no papers $(y,u)=(0,0)$. Recruiters receive the signal $y$ from an applicant's resume. In addition, they conduct a hiring test (interview), which is meant to sort applicants based on their understanding. However, this test is prone to false positives. Individuals with no understanding (i.e., $u=0$) fail the test ($t=\text{F}$) with probability $\tau \in (0,1)$ and pass it ($t=\text{P}$) with probability $1-\tau > 0$. The recruiter makes a decision based on the joint variable $(y,t)$ coming from the candidate's resume and interview performance. If hired, an applicant receives $W-c$, and if rejected, $-c$, where $c=\delta_\theta$ after deep study and $c=0$ after shallow study. If an applicant is hired, the recruiter gets payoff $Ay + Bu$ with $A,B>0$ representing how much the recruiter values the productivity and the understanding of their applicant, respectively. If an applicant is rejected, the recruiter earns an opting-out payoff $r>0$.}
    \label{fig:game-tree}
\end{figure}

\noindent In addition to viewing their resume, the interviewing committee performs an \emph{understanding test} on the candidate (e.g., an interview). The result of this test is either $t=\text{P}$ (for `pass') or $t=\text{F}$ (for `fail'). An applicant with a rich (deep) understanding of the field has $u=1$, and an applicant with partial (shallow) understanding of the field has $u=0$. A $u=1$ candidate passes the test with probability 1. A $u=0$ candidate fails the test (i.e., $t=\text{F}$) with probability $\tau$ and passes it (i.e., $t=\text{P}$) with probability $1-\tau$. The variable $\tau \in (0,1)$ sets the accuracy of the test. At $\tau = 1$, the test perfectly sorts candidates based on understanding. At $\tau = 0$, the test is useless and everyone passes. Mathematically,

\[\mathbb{P}(t=\text{P}\:|\:u=1) = 1 \quad \text{and }\quad \mathbb{P}(t=\text{F}\:|\:u=0) = \tau. \]\vspace{0.1cm}

\noindent We exclude the $\tau\in\{0,1\}$ boundary cases.\\

\noindent \textbf{Recruiter payoff.} The interviewer decides whether to hire or to reject based on the joint outcome $(y,t)$. If they accept a candidate, their payoff is $Ay + Bu$, where $y$ and $u$ are the paper and understanding values of the hired candidate. If they reject a candidate, they get an opting-out payoff $r$. Parameters $A$ and $B$ respectively capture how much the recruiter values papers and understanding in their candidate, independently. \\

\noindent \textbf{Applicant type and effort strategy.} In this model, there are three kinds of applicants: \emph{(i)} high potential, $H$, \emph{(ii)} medium potential, $M$, and \emph{(iii)} low potential, $L$. Only the first two types choose a strategy. Think of these strategies as pre-interview career trajectories. There are two strategies: \emph{deep} and \emph{shallow + AI}. A deep trajectory guarantees full understanding, i.e., $u=1$, whereas a shallow trajectory yields partial or no understanding of the field, i.e., $u=0$.
When it comes to papers, a high-potential individual produces papers regardless of trajectory choice ($y=1$). A medium-potential individual, however, only produces papers with `shallow + AI'. (In a model extension, we study the case where a high-potential individual produces \emph{many more} papers than a medium-potential individual when going the shallow + AI route. We do this by making the variable $y$ a ternary variable with $y$ taking values $0,1,$ and $2$.) The third and final applicant type (low-potential) is non-strategic. It has no understanding, $u=0$, and few (or no) papers, $y=0$. The population shares of these types are $\pi_H$, $\pi_M$, and $\pi_L$, respectively (each $>0$), and $\pi_H + \pi_M + \pi_L = 1$.\\

\noindent \textbf{Applicant payoff.} Getting hired gives an applicant a payoff $W>0$.
Deep study costs a high-potential applicant $\delta_H$ more than shallow study, and a medium-potential applicant $\delta_M$ more, with $0<\delta_H<\delta_M<W$.
Only this difference matters, so we normalize the cost of shallow study to zero.
An applicant therefore pays $c=\delta_\theta$ after deep study and $c=0$ after shallow study, and receives $W-c$ if hired and $-c$ if rejected.\\

\begin{table}[tp!]
\centering
\small
\renewcommand{\arraystretch}{1.3}
\setlength{\tabcolsep}{5pt}
\begin{tabularx}{\textwidth}{@{}>{\centering\arraybackslash}p{2.0cm} X >{\centering\arraybackslash}p{1.9cm} >{\centering\arraybackslash}p{1.9cm}@{}}
\toprule
Symbol & Meaning and values & Applicant knows? & Recruiter knows? \\
\midrule
\multicolumn{4}{@{}l}{\textit{Applicant type (drawn by Nature)}}\\
$\theta$ & high ($H$), medium ($M$), or low ($L$) potential & \yes\ (private) & \no \\
$\pi_H,\ \pi_M,\ \pi_L$ & Population shares of types, $\pi_H+\pi_M+\pi_L=1;\: \pi_\theta > 0$ & \yes & \yes \\
\addlinespace \addlinespace
\multicolumn{4}{@{}l}{\textit{Applicant's choice}}\\
$a$ & Trajectory chosen by $H$ and $M$: &&\\&deep ($\D$) or shallow+AI ($\Sh$); $L$ does not choose & \yes\ (own) & \no \\
$u$ & Understanding: $u=1$ after $\D$, $u=0$ after $\Sh$ or for $L$ & \yes\ (own) & \no \\
$\delta_H$,\ $\delta_M$ & Net cost of choosing deep over shallow, $0\!<\delta_H\!<\delta_M\!<W$ & \yes & \yes \\
\addlinespace \addlinespace
\multicolumn{4}{@{}l}{\textit{What the recruiter observes}}\\
$y$ & Output in papers: &&\\
&(Baseline model): $y=1$, many papers; $y=0$, few papers.&&\\
&(Extension): $y=2$, super; $y=1$, many; $y=0$, few papers & \yes & \yes \\
$t$ & Test outcome: pass ($\Pass$) or fail ($\Fail$) & \yes & \yes \\
$\tau$ & Test accuracy: probability that an applicant without understanding fails; applicants with understanding always pass & \yes & \yes \\
\addlinespace \addlinespace
\multicolumn{4}{@{}l}{\textit{Payoffs}}\\
$W$ & Value of being hired to the applicant & \yes & \yes \\
$A$ & Value to the recruiter of each unit of output $y$ & \yes & \yes \\
$B$ & Value to the recruiter of a hire who understands ($u=1$) & \yes & \yes \\
$r$ & Recruiter's outside option (value of rejecting) & \yes & \yes \\
\bottomrule
\end{tabularx}
\caption{\textbf{Elements of the effort-and-hiring game}. This table summarizes the variables and parameters in the effort-and-hiring game described in Section \ref{sec:game}. The applicant's type $\theta$, trajectory $a$, and understanding $u$ are private to the applicant. The recruiter observes only output $y$ and the test outcome $t$ (which may produce false positives by passing $u=0$ players with probability $1-\tau>0$). All other parameters are common knowledge. A high-potential applicant produces papers $y=1$ regardless of trajectory, but medium-potential applicants only get $y=1$ with shallow+AI. If hired, the applicant receives $W-c$ and the recruiter gets $Ay+Bu$; if rejected, the applicant receives $-c$ and the recruiter, $r$, where $c=\delta_\theta$ after deep work and $c=0$ otherwise. In an extended model, we allow $y$ to take three values, $y\in\{0,1,2\}$, representing different levels of research output. With AI, a high-potential individual achieves $y=2$ (called a super-output), whereas a medium-potential individual achieves $y=1$. Without AI, a high-potential individual also achieves $y=1$.}
\label{tab:game-description}
\end{table}

\noindent \textbf{Recruiter strategy}. The recruiter employs a strategy $\rho(y,t) \in [0,1]$, the probability to accept an applicant with a resume worth $y$ and a test result $t$. The variable $y\in\{0,1\}$ (in the baseline model) and $y \in \{0,1,2\}$ (in the extended model). The test result $t\in\{\Pass,\Fail\}$. We highlight that a recruiter never sees the true type of the applicant. Nor do they know an applicant's understanding-level perfectly. They only get noisy signals of these through $y$ and $t$. \\

\noindent We show a schematic figure of this game in \textbf{Figure} \ref{fig:game-tree}. In \textbf{Table} \ref{tab:game-description}, we summarize our model parameters and highlight which parameters are known to which players. \\

\section{Equilibrium analysis of the effort-and-hiring game}
\label{sec:equilibrium}
Formally, our setup in Section~\ref{sec:game} is called a sequential game with incomplete information. The relevant game-theoretic equilibrium notion for this class of games is the \emph{perfect Bayesian equilibrium}.\\

\noindent What does this particular equilibrium notion mean? In plain terms, it represents a stable pattern of behavior on both sides of the game together with a consistent way of reading the evidence at hand. It has three ingredients: (a) each type of applicant chooses an effort trajectory that serves it best, given how recruiters respond; (b) the recruiter accepts or rejects each kind of applicant in the way that serves the recruiter best, given what it believes about that applicant; (c) these beliefs are \emph{correct on average}: they follow from the share of each type in the population and from what each type actually does.\\

\noindent For example, if in equilibrium only high-potential applicants work deeply, then a recruiter who sees many papers and a passed test knows that this applicant is either a high-potential deep worker or a medium-potential applicant who used AI and passed the test by luck. In equilibrium, the recruiter weighs the two possibilities by the likelihood of their outcome. At equilibrium, neither side has the payoff-incentive to change their (work or hiring) behavior unilaterally. In this paper, we characterize the pure-strategy perfect Bayesian equilibria of this game. Later, we also perform evolutionary simulations to study the behaviors that are learned when multiple equilibria exist simultaneously.

\subsection*{The five pure-strategy equilibria of the baseline model}

\textbf{Baseline-model equilibria.} In the baseline model, one of five behavioral patterns is stable at a pure-strategy perfect Bayesian equilibrium. \textbf{Table} \ref{tab:equilibria}B1 summarizes these five equilibria, specifying each party's behavior, the conditions for equilibrium existence, and the recruiter's beliefs at equilibria. To aid interpretation, \textbf{Table} \ref{tab:equilibria}A provides the key terms and definitions. The five equilibrium patterns of the baseline model, labeled $\mathbf{E}_1$--$\mathbf{E}_5$, are described below. \\
\begin{enumerate}
    \item $\mathbf{E}_1$: All strategic applicants choose shallow; recruiter hires nobody.
    \item $\mathbf{E}_2$: Only $H$ applicants choose deep; recruiters hire only if they see $(y,t)  = (1,\text{P})$.
    \item $\mathbf{E}_3$: All strategic applicants choose deep; recruiter only hires if they see $t = \text{P}$.
    \item $\mathbf{E}_4$: $H$ applicants go shallow, $M$ applicants go deep; recruiter only hires $(y,t) = (0,\text{P})$.
    \item $\mathbf{E}_5$: All strategic applicants choose shallow; all with $y=1$ hired.
\end{enumerate}
\vspace{0.5cm}
\noindent The equilibrium $\mathbf{E}_5$, wherein strategic applicants choose the shallow route and the recruiter hires everyone with papers (i.e., $y=1$, or basically every strategic applicant), is trivial. It only appears when $A \geqslant r$---that is, when the recruiter values a many-paper applicant more than their opting-out payoff. In fact, in this parameter regime, this is the only equilibrium because output alone already makes a hire worthwhile, which empties the test of any consequence. Applicants too find no reason to invest in costly understanding.\\


\begin{table}
\centering
\footnotesize
\renewcommand{\arraystretch}{1.08}
\setlength{\tabcolsep}{4pt}

\begin{tabularx}{\textwidth}{@{}>{\raggedright\arraybackslash}p{4.1cm} X@{}}
\multicolumn{2}{@{}l}{\textbf{Panel A. Notation}}\\
\toprule
Symbol & Definition \\
\midrule
$\sigma_H,\ \sigma_M \in \{0,1\}$ & Effort strategy of high-potential ($H$) and medium-potential ($M$) applicants:\\& deep work ($=1$) or shallow work with AI ($=0$).\\
$\rho_{1\Pass},\rho_{1\Fail},\rho_{0\Pass},\rho_{0\Fail}$ & Recruiter strategy; probability to accept after observing $y\in\{0,1\}$ and $t \in \{\Pass,\Fail\}$ \\
$\rho_2$ & Probability of accepting super-output, $y=2$ (in extended model only). \\
$\mu_1,\ \mu_0$ & Recruiter belief that an applicant who passed the test with many papers ($\mu_1$) or few papers ($\mu_0$) truly understands. \\
$\bar\mu_1=\pi_H/[\pi_H+(1-\tau)\pi_M]$ & Value of $\mu_1$ when only high-potential applicants work deeply. \\
$\bar{\mu}_0=\pi_M/[\pi_M+(1-\tau)\pi_L]$ & Value of $\mu_0$ when medium-potential applicants work deeply. \\
$\mathcal{B}_1=(r-A)/\bar\mu_1,\ \mathcal{B}_0=r/\bar{\mu}_0$ & Value of understanding above which the recruiter hires applicants who passed with many ($\mathcal{B}_1$) or few ($\mathcal{B}_0$) papers. \\
\bottomrule
\end{tabularx}

\vspace{0.8em}

\begin{tabularx}{\textwidth}{@{}>{\raggedright\arraybackslash}p{3.45cm} c c c >{\raggedright\arraybackslash}X@{}}
\multicolumn{5}{@{}l}{\textbf{Panel B. Pure-strategy perfect Bayesian equilibria}}\\
\toprule
Equilibrium & \begin{tabular}[c]{@{}c@{}}Applicant\\ strategy\\ $(\sigma_H,\sigma_M)$\end{tabular}
 & \begin{tabular}[c]{@{}c@{}}Recruiter strategy\\ $(\rho_{1\Pass},\rho_{1\Fail},\rho_{0\Pass},\rho_{0\Fail};\ \rho_2)$\end{tabular}
 & \begin{tabular}[c]{@{}c@{}}Recruiter\\ belief\\ $(\mu_1,\mu_0)$\end{tabular}
 & Exists if and only if \\
\midrule
\multicolumn{5}{@{}l}{\textit{B1. Baseline (a high-potential applicant using AI produces $y=1$)}}\\
\addlinespace[2pt]
\multicolumn{5}{@{}l}{\quad Depth regime, $A<r$}\\
$\mathbf{E}_1$: all shallow, no hires & $(0,0)$ & $(0,0,0,0;\ \text{--})$ & $(0,0)$ & always \\
$\mathbf{E}_2$: only $H$ deep & $(1,0)$ & $(1,0,0,0;\ \text{--})$ & $(\bar\mu_1,0)$ & $B\geqslant\mathcal{B}_1$ and $\tau W\geqslant\delta_H$ \\
$\mathbf{E}_3$: all deep & $(1,1)$ & $(1,0^\dagger,1,0;\ \text{--})$ & $(1,\bar{\mu}_0)$ & \mbox{$B\geqslant\mathcal{B}_0$ and $\tau W\geqslant\max(\delta_H,\delta_M)$} \\
$\mathbf{E}_4$: reverse sorting & $(0,1)$ & $(0,0,1,0;\ \text{--})$ & $(0,\bar{\mu}_0)$ & $B\geqslant\mathcal{B}_0$ \\
\addlinespace[2pt]
\multicolumn{5}{@{}l}{\quad Output regime, $A\geqslant r$}\\
$\mathbf{E}_5$: shallow, $y=1$ hired & $(0,0)$ & $(1,1,0,0;\ \text{--})$ & $(0,0)$ & always; unique \\
\midrule
\multicolumn{5}{@{}l}{\textit{B2. Extension (a high-potential applicant using AI produces super-output $y=2$)}}\\
\addlinespace[2pt]
\multicolumn{5}{@{}l}{\quad $A<r/2$}\\
$\mathbf{E}_1$: shallow, no hires & $(0,0)$ & $(0,0,0,0;\ 0)$ & $(0,0)$ & always \\
$\mathbf{E}_2$: $H$ deep & $(1,0)$ & $(1,0,0,0;\ 0^\dagger)$ & $(\bar\mu_1,0)$ & $B\geqslant\mathcal{B}_1$ \\
$\mathbf{E}_3$: all deep & $(1,1)$ & $(1,0^\dagger,1,0;\ 0^\dagger)$ & $(1,\bar{\mu}_0)$ & $B\geqslant\mathcal{B}_0$ and $\tau W\geqslant\delta_M$ \\
$\mathbf{E}_4$: reverse sorting & $(0,1)$ & $(0^\dagger,0^\dagger,1,0;\ 0)$ & $\bigl(<\!\tfrac{r-A}{B},\,\bar{\mu}_0\bigr)$ & $B\geqslant\mathcal{B}_0$\\
\addlinespace[2pt]
\multicolumn{5}{@{}l}{\quad $r/2\leqslant A<r$}\\
$\mathbf{E}_6$: shallow, $H$ hired & $(0,0)$ & $(0,0,0,0;\ 1)$ & $(0,0)$ & always \\
$\mathbf{E}_7$: reverse sorting & $(0,1)$ & $(0^\dagger,0^\dagger,1,0;\ 1)$ & $(<\frac{r-A}{B},\bar{\mu}_0)$ & $B\geqslant\mathcal{B}_0$ \\
& $(0,1)$ & $(1^\dagger,0^\dagger,1,0;\ 1)$ & $(\geqslant\frac{r-A}{B},\bar{\mu}_0)$ & $B\geqslant\mathcal{B}_0;\: \tau W \geqslant \delta_M$\\
\addlinespace[2pt]
\multicolumn{5}{@{}l}{\quad Output regime, $A\geqslant r$}\\
$\mathbf{E}_5$: shallow, $y \geqslant 1$ hired & $(0,0)$ & $(1,1,0,0;\ 1)$ & $(0,0)$ & always; unique \\
\bottomrule
\end{tabularx}

\caption{\textbf{Pure-strategy perfect Bayesian equilibria of the effort-and-hiring game.}
Panel~\textbf{A} defines notation.
Panel~\textbf{B} lists every pure-strategy equilibrium outcome for $0\!<\!\tau\!<1$ in the baseline (B1) and super-output (B2) models; outcomes not listed in a regime do not exist there.
Each row gives the conditions under which its supporting profile is an ordinary perfect Bayesian equilibrium. Outcomes with several supporting profiles, such as $\mathbf{E}_7$, have one row per profile. We assume $0<\delta_H<\delta_M<W$ and break ties in favor of acceptance.
A dash (--) marks an observation impossible in the baseline.
A dagger ($^\dagger$) in the strategy profile marks a decision after an observation that never occurs in equilibrium. These actions rest on some off-equilibrium beliefs. We discuss them in Section~\ref{sec:equilibrium}.}

\label{tab:equilibria}
\end{table}


\noindent When output alone does not justify a hire ($A<r$), the recruiter accepts an applicant only if it believes the applicant understands, so every equilibrium rests on what a passed test is believed to reveal. In $\mathbf{E}_1$, nobody works deeply, so a pass reveals nothing ($\mu_1=\mu_0=0$) and the recruiter rejects everyone. An applicant who deviates to deep work cannot be told apart from lucky AI users. They bear the cost $\delta_\theta$ for nothing, and thus the trap sustains itself. In $\mathbf{E}_2$, only high-potential applicants work deeply. Applicants with many papers who pass are then genuine understanders with probability $\bar\mu_1$ (from the perspective of recruiters), and the recruiter hires them provided understanding is valuable enough ($B\geqslant\mathcal{B}_1$). High-potential applicants find depth worthwhile when the test catches enough AI users to make the job premium $W$ cover the cost ($\tau W\geqslant\delta_H$). \\

\noindent In $\mathbf{E}_3$, medium-potential applicants also work deeply, accepting few papers in exchange for being recognized as competent. This removes the AI users from the many-paper pool ($\mu_1=1$), and it requires the recruiter to hire few-paper passers ($B\geqslant\mathcal{B}_0$), who are mostly genuine but include low-potential applicants who passed by luck. \\

\noindent $\mathbf{E}_4$ is the mirror image of $\mathbf{E}_2$: only medium-potential applicants work deeply and are hired through the few-paper-and-pass channel. Every many-paper applicant who passes is a lucky AI user. Since the recruiter only hires those that have few papers but also pass the test, high-potential candidates are not tempted to try the costly route of going deep. This is because there is no way for them to escape their inherent productivity (remember that high-potential applicants produce $y=1$ regardless of effort). As a result, in this equilibrium, the high-potential applicants end up shallow and unemployed. The four listed pure-strategy equilibria for $A < r$ may simultaneously exist depending on the parameter values (\textbf{Table} \ref{tab:equilibria}, Figure \ref{fig:equilibrium-regions}A). In a later section, we discuss the evolutionary selection among these various equilibria.

\subsection*{Extra equilibria in the extended super-output model}

In the extended model, we consider three levels of paper output, $y=0,1,2$. The $y=2$ case represents the \textit{super-output} which is achieved by high-potential applicants using AI. When high-potential applicants go deep or when medium-potential applicants go shallow, they produce $y=1$. We call $y=0,1, 2$, few papers, many papers, and super-output, respectively. Because super-output can only come from a high-potential AI user, it perfectly reveals that the applicant lacks understanding and is worth $2A$ to the recruiter, regardless of the test. Whether the recruiter hires them therefore depends only on whether $2A \geqslant r$, and this splits the parameter space into three regimes.\\

\begin{figure}
    \centering
    \includegraphics[width=\linewidth]{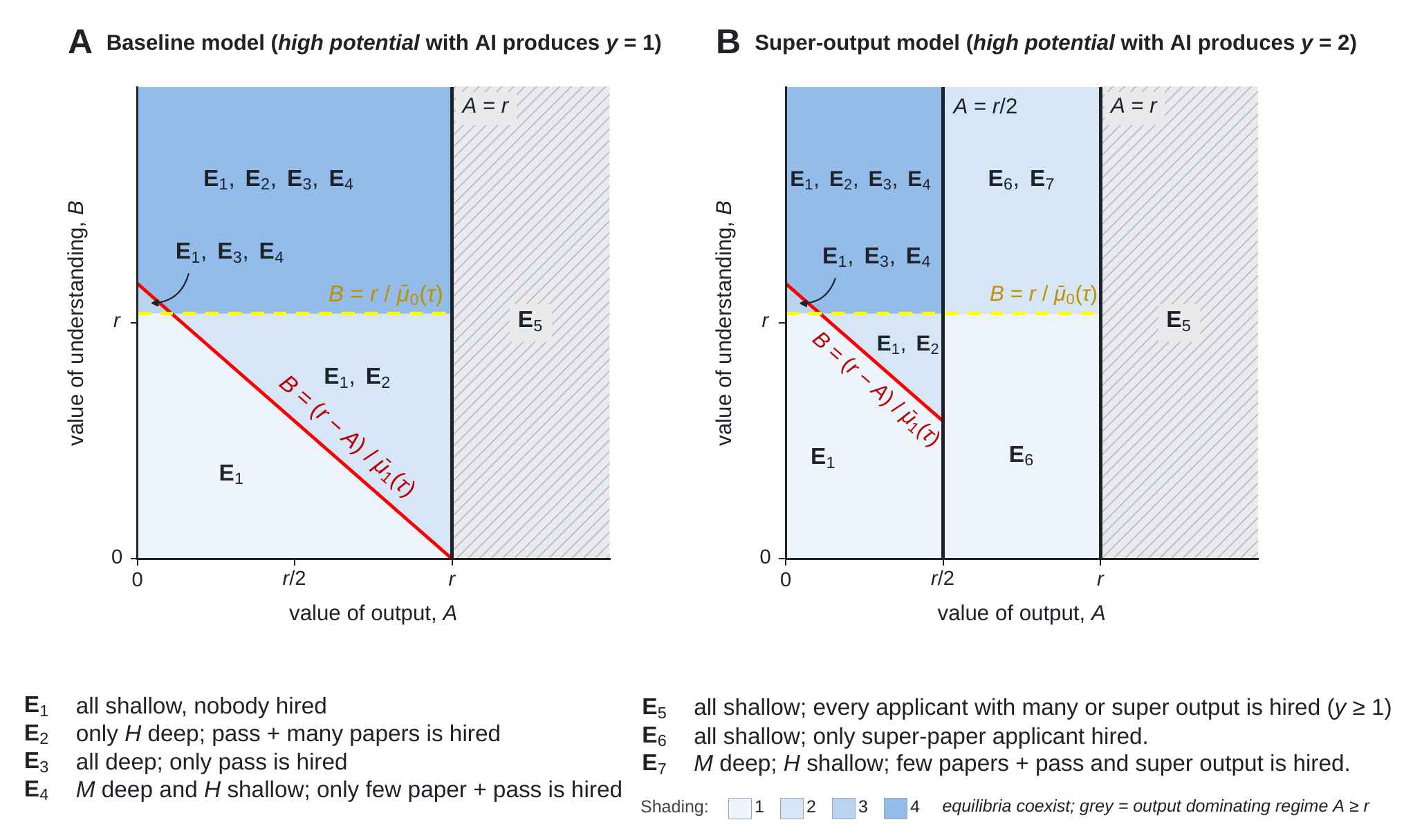}
    \vspace{0.2cm}
    \caption{\textbf{Existence regions of pure-strategy perfect Bayesian equilibria.}
    Each region of the $A$--$B$ plane is labeled with the perfect Bayesian equilibrium outcomes that exist there, for the baseline model (\textbf{A}) and the super-output model (\textbf{B}).
    The red line is $B=\mathcal{B}_1$, the dashed line is $B=\mathcal{B}_0$, and the vertical lines mark $A=r$ and, in \textbf{B}, $A=r/2$. These are all defined in Table~\ref{tab:equilibria}A.
    As we see, several equilibrium outcomes can coexist for a particular parameter combination. The shade of blue in the parameter plane gives the number of distinct equilibrium outcomes for the underlying parameter (not strategy profiles).
    Parameters are $(\pi_H,\pi_M,\pi_L)=(0.3,0.5,0.2)$, $\tau=0.9$, $\delta_H=0.2$, $\delta_M=0.4$, and $W=r=1$.}
    \label{fig:equilibrium-regions}
\end{figure}

\noindent When $A<r/2$, super-output is not worth hiring, so the high-potential applicants can no longer hide behind AI. A high-potential applicant who chooses shallow study is rejected outright, and the same four equilibria $\mathbf{E}_1$--$\mathbf{E}_4$ arise as in the baseline. However, depth is now easier to sustain: high-potential applicants go deep whenever the recruiter trusts many-paper passers ($B \geqslant \mathcal{B}_1$), irrespective of the test discernment $\tau$. Reverse sorting $\mathbf{E}_4$ still exists, but only because the recruiter holds a pessimistic belief about an observation that no one produces (i.e., $y=1$). Under the D1 refinement, however, this belief is implausible, and $\mathbf{E}_4$ is ruled out when depth is cheap enough for high-potential applicants, $\delta_H(1-\tau) < W - \delta_M$. For more details on the D1 refinement, see the note on off-equilibrium beliefs later in this section.\\

\noindent When $r/2 \leqslant A < r$, super-output alone justifies a hire, and two new pure-strategy equilibria replace $\mathbf{E}_1$--$\mathbf{E}_4$, both of which have the property that high-potential applicants do not invest in understanding because AI now guarantees a job at no cost. In $\mathbf{E}_6$, medium-potential applicants also stay shallow; their many papers are not worth hiring, so only the high-potential applicants are hired, on the strength of super-output. In $\mathbf{E}_7$, medium-potential applicants instead work deeply and are hired through the few-paper channel ($B \geqslant \mathcal{B}_0$), while the high-potential applicants are hired for their super-output. In both equilibria, understanding, if present at all, is concentrated among medium-potential applicants.\\

\noindent Finally, when $A \geqslant r$, the output-dominant regime of the baseline model reappears. Every applicant with $y \geqslant 1$ is hired regardless of the test, nobody works deeply, and $\mathbf{E}_5$ is the unique equilibrium. The super-output extension thus moves the threshold beyond which high-potential applicants abandon understanding from $A=r$ down to $A=r/2$. In other words, the more productive an AI makes the high-potential applicants, the lower
    the recruiter's per-unit valuation of output needs to be before depth disappears among them. The equilibrium regions for the super-output model are summarized in \textbf{Figure} \ref{fig:equilibrium-regions}B.\\

\noindent \textbf{What is the likelihood that hired candidates possess understanding at each equilibrium?} Equilibria differ not only in who goes deep but also in the outcomes the recruiter cares about.
We therefore compare them on two measures: the share of all applicants who are hired and the hiring quality defined as the probability $\Pr(u=1 \mid \text{hired})$ that a hired candidate understands the subject. This is summarized in \textbf{Table}~\ref{tab:hire-understand}. \\

\begin{table}[h!]
\centering
\footnotesize
\setlength{\tabcolsep}{4pt}
\renewcommand{\arraystretch}{1.6}
\vspace{6pt}
\begin{tabular}{@{}l l l c c @{}}
\toprule
Eq. & Who is hired & Hire rate & $\mathbb{P}(u=1\mid\text{hired})$ & Example \\
\midrule
\multicolumn{5}{@{}l}{\textit{Panel A: Baseline model}}\\

$\mathbf{E}_1$ & nobody & $0$ & -- & -- \\
$\mathbf{E}_2$ & H (deep); M (shallow) who pass & $\pi_H\!+\!(1-\tau)\pi_M$ & $\dfrac{\pi_H}{\pi_H\!+\!(1-\tau)\pi_M}$ & 0.857 \\[10pt]
$\mathbf{E}_3$ & H, M (deep); L who pass & $\pi_H\!+\!\pi_M\!+\!(1-\tau)\pi_L$ & $\dfrac{\pi_H\!+\!\pi_M}{\pi_H\!+\!\pi_M\!+\!(1-\tau)\pi_L}$ & 0.976 \\[10pt]
$\mathbf{E}_4$ & M (deep); L who pass & $\pi_M\!+\!(1-\tau)\pi_L$ & $\dfrac{\pi_M}{\pi_M\!+\!(1-\tau)\pi_L}$ & 0.962 \\
$\mathbf{E}_5$ & H, M (shallow) & $\pi_H\!+\!\pi_M$ & $0$ & 0 \\
\midrule
\multicolumn{5}{@{}l}{\textit{Panel B: Super-output model}}\\
$\mathbf{E}_1$ & nobody & $0$ & -- & -- \\
$\mathbf{E}_2$ & H (deep); M (shallow) who pass & $\pi_H\!+\!(1-\tau)\pi_M$ & $\dfrac{\pi_H}{\pi_H\!+\!(1-\tau)\pi_M}$ & 0.857 \\[10pt]
$\mathbf{E}_3$ & H, M (deep); L who pass & $\pi_H\!+\!\pi_M\!+\!(1-\tau)\pi_L$ & $\dfrac{\pi_H\!+\!\pi_M}{\pi_H\!+\!\pi_M\!+\!(1-\tau)\pi_L}$ & 0.976 \\[10pt]
$\mathbf{E}_4$ & M (deep); L who pass & $\pi_M\!+\!(1-\tau)\pi_L$ & $\dfrac{\pi_M}{\pi_M\!+\!(1-\tau)\pi_L}$ & 0.962 \\[10pt]
$\mathbf{E}_6$ & H (shallow, $y=2$) & $\pi_H$ & $0$ & 0 \\
$\mathbf{E}_7$ & \begin{tabular}[c]{@{}l@{}}H (shallow, $y=2$); M (deep);\\ L who pass\end{tabular} & $\pi_H\!+\!\pi_M\!+\!(1-\tau)\pi_L$ & $\dfrac{\pi_M}{\pi_H\!+\!\pi_M\!+\!(1-\tau)\pi_L}$ & 0.610 \\
$\mathbf{E}_5$ & H, M (shallow) & $\pi_H\!+\!\pi_M$ & $0$ & 0 \\
\bottomrule
\end{tabular}
\caption{\onehalfspacing \textbf{Rate of hiring and the probability that a hired candidate understands in each listed pure-strategy equilibrium of the two models.} Panel \textbf{A} reports the baseline results, and panel \textbf{B} the extended results. The last column gives values for $\pi_H=0.3$, $\pi_M=0.5$, $\pi_L=0.2$, and $\tau=0.9$.}
\label{tab:hire-understand}
\end{table}

\noindent The share of hires who understand the subject is highest in either $\mathbf{E}_2$ or $\mathbf{E}_3$. Between these two, which one has the higher share depends only on the type distribution $\pi$:
\[
\Pr(u=1\mid\text{hired})\big|_{\mathbf{E}_3}\;\geqslant\;\Pr(u=1\mid\text{hired})\big|_{\mathbf{E}_2}
\quad\Longleftrightarrow\quad
\pi_H\,\pi_L\;\leqslant\;\pi_M\,(\pi_H+\pi_M).
\]
The two equilibria make different mistakes.
In $\mathbf{E}_2$, the errors are $M$ applicants who go shallow and slip past the test.
In $\mathbf{E}_3$, the errors are low-potential $L$ applicants who slip past it.
$\mathbf{E}_3$ therefore wins when low-potential applicants are scarce relative to medium-potential ones, and $\mathbf{E}_2$ wins when they are common, because $\mathbf{E}_2$ screens them out entirely.
The test accuracy $\tau$ scales both kinds of error by the same factor $1-\tau$, so it changes how large the errors are but not which equilibrium does better.\\

\noindent\textbf{Note on off-equilibrium-path beliefs in our equilibria characterization.}
In an equilibrium, some observations never occur. For example, if every applicant works deeply (assuming that there are no low-potential applicants, who do not have the strategic option of working deeply), the recruiter never sees anyone fail the test. Because such an observation has probability zero, Bayes' rule cannot tell the recruiter what to believe if it were to happen. These are called \emph{off-equilibrium-path beliefs}. They still matter, because they determine how the recruiter would respond to a deviation, and therefore whether anyone wants to deviate. We discipline them with two standard refinements.
\emph{Consistency} requires that if only one kind of applicant could possibly have produced an observation, the recruiter attributes it to that kind.
The \emph{D1 criterion} applies when several kinds of applicant could have produced an unexpected observation. It requires the recruiter to attribute the observation to the kind that would gain from it under the widest range of recruiter responses, that is, to the applicant with the strongest incentive to deviate.
In \textbf{Table}~\ref{tab:equilibria}, a dagger ($^\dagger$) marks a recruiter decision that rests on such a belief. The three cases are:\\

\smallskip
\noindent (i)~\emph{Failed test with many papers} ($\rho_{1\Fail}=0^\dagger$ in $\mathbf{E}_3$ of both models, and in $\mathbf{E}_4$ and $\mathbf{E}_7$ of the extended model).
Only an applicant without understanding can fail the test, so consistency forces the belief $u=0$. Since output alone is not worth hiring ($A<r$), the recruiter rejects.\\

\noindent (ii)~\emph{Super-output} ($\rho_2=0^\dagger$ in $\mathbf{E}_2$ and $\mathbf{E}_3$ of the extended model, $A<r/2$).
Only a high-potential applicant using AI can produce $y=2$, so consistency again forces $u=0$. Since super-output alone is not worth hiring ($2A<r$), the recruiter rejects.\\

\noindent (iii)~\emph{Many papers with a passed test} ($\rho_{1\Pass}=0^\dagger$ in $\mathbf{E}_4$; $\rho_{1\Pass}=0/1^\dagger$ in $\mathbf{E}_7$ of the extended model).
In these equilibria, no applicant produces $y=1$, since high-potential applicants produce super-output and medium-potential applicants produce few papers.
Two deviators could produce this observation: a high-potential applicant who works deeply, and a medium-potential applicant who uses AI.
Consistency alone therefore does not pin down the belief $\mu_1$.
Any $\mu_1<(r-A)/B$ supports rejection, and any $\mu_1\geqslant(r-A)/B$ supports acceptance.
D1 attributes the observation to the type that would gain from deviating for the wider range of recruiter responses.
A high-potential applicant gains whenever the probability of acceptance after this observation exceeds $\delta_H/W$, provided they are not already hired.
A medium-potential applicant gains only when it exceeds $(W-\delta_M)/(W(1-\tau))$, which is impossible when $\delta_M\leqslant\tau W$.\\

\noindent \emph{(a)} In $\mathbf{E}_4$ ($A<r/2$), high-potential applicants are rejected for their super-output and may gain from working deeply.
If $\delta_H(1-\tau)<W-\delta_M$, D1 attributes the observation to them ($\mu_1=1$).
The recruiter would then hire, high-potential applicants would deviate, and $\mathbf{E}_4$ is eliminated.
If $\delta_H(1-\tau)\geqslant W-\delta_M$, D1 does not single out the high-potential type, and $\mathbf{E}_4$ survives.
At equality, the two thresholds coincide, so D1 cannot single out either type and the equilibrium survives.\\

\noindent \emph{(b)} In $\mathbf{E}_7$ ($r/2\leqslant A<r$), high-potential applicants are already hired for their super-output and can never gain by working deeply.
If $\delta_M>\tau W$, only a medium-potential AI user could gain, so D1 sets $\mu_1=0$ and rejection is the only supported response.
Acceptance would in any case not be an equilibrium response here, since medium-potential applicants would then switch to shallow study.
If $\delta_M\leqslant\tau W$, neither type can gain from producing this observation, so D1 places no restriction on beliefs.
Both acceptance and rejection are then consistent with D1, which is why the table lists $\rho_{1\Pass}=0/1^\dagger$.
In every case, $\mathbf{E}_7$ survives, and the off-path response does not affect who studies deeply or who is hired.\\

\noindent Every other entry in \textbf{Table}~\ref{tab:equilibria} concerns observations that occur on the equilibrium path, and the corresponding beliefs follow directly from Bayes' rule.

\section{Selection between simultaneously existing equilibria}\label{sec:learning}

Our equilibrium analysis shows that multiple equilibria can simultaneously exist, depending on the game parameters (\textbf{Figure} \ref{fig:equilibrium-regions}). To investigate which equilibria are \textit{attainable}, we study a learning process in populations of applicants and recruiters. Rather than assuming that both parties coordinate on a particular equilibrium, we allow their strategies to evolve through a myopic learning process, where individuals copy strategies from others in the population if those other strategies currently yield a higher payoff than their own strategy.\\

\noindent To this end, we consider three populations: one for high-potential applicants, one for medium-potential applicants, and one for recruiters. Each population is of size $N$. Applicants choose between D and S (for deep and shallow). Recruiters choose one of 16 pure strategies $(\rho_{1\Pass},\rho_{1\Fail},\rho_{0\Pass},\rho_{0\Fail}) \in \{0,1\}^4$ in the baseline model and among 32 in the extended model with $\rho_2 \in \{0,1\}$ included in the strategy. \\

\noindent We study the imitation-and-exploration dynamic in these populations with exploration happening very rarely \citep{fudenberg2006imitation}. We assume that individuals learn strategies from within their population (i.e., a high-potential applicant never imitates a medium-potential applicant and \emph{vice versa}). We set the applicant-pool distribution $(\pi_H,\pi_M,\pi_L)$, the type-sensitive costs $\delta_H$ and $\delta_M$, the job premium $W$, and the test accuracy $\tau$, exogenously for each simulation. We use our simulations to quantify how often, in the long run, high-potential and medium-potential applicants choose deep versus shallow strategies, and how often recruiters adopt each of their 16 (or 32) strategies. The simulations thus reveal behavior that emerges through imitation and exploration in both applicants and recruiters, rather than through coordination on a particular equilibrium. Full details of the learning process are provided in \textbf{Methods}.\\

\begin{figure}
    \centering
    \includegraphics[width=0.65\linewidth]{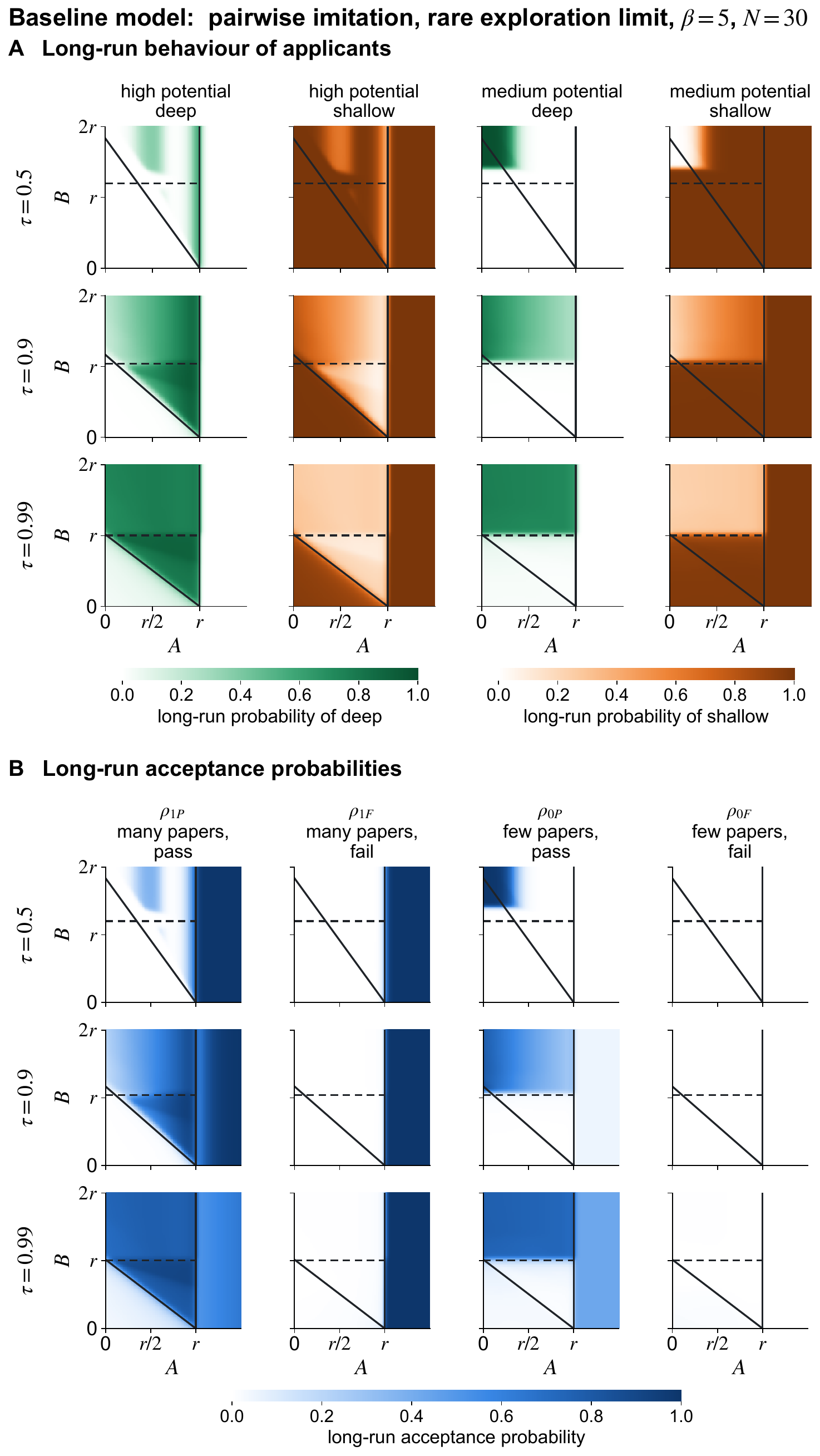}
    \caption{\textbf{Long-run behavior of applicants and recruiters in a learning process (baseline model).}  
    We study how strategies evolve in populations of high-potential applicants, medium-potential applicants, and recruiters as a result of imitation and rare exploration.
    For full details of the learning process, see \textbf{Methods}. In panel \textbf{A}, we plot the long-run likelihood that the high-potential and medium-potential populations use a deep (or shallow) trajectory. In panel \textbf{B}, we plot the long-run likelihood that recruiters accept after one of four possible outcomes of an evaluation $(y,t) \in \{(1,\Pass),(1,\Fail),(0,\Pass),(0,\Fail)\}$. We plot this as we vary the parameters $A$ and $B$ in the horizontal and vertical axes, respectively, and the accuracy of the test, $\tau$, in rows. Game parameters are the same as those in \textbf{Figure} \ref{fig:equilibrium-regions}. The population size is set at $N=30$ and the strength of imitation, at $\beta = 5$.}
    \label{fig:baseline-evolution}
\end{figure}

\noindent Let us first discuss the learning outcomes in \textbf{Figure} \ref{fig:baseline-evolution}, which shows the long-run behavior of applicants and recruiters in the baseline model. In the baseline model, test accuracy does not matter once output alone justifies hiring ($A \geqslant r$).
There, both strategic types learn to go shallow, and recruiters learn to hire every applicant with many papers, whether or not they pass the test ($\rho_{1P}, \rho_{1F} \approx 1$).
When the test is accurate, recruiters also occasionally hire few-paper applicants who pass.
These are only the low-potential applicants who get through by luck, so accepting them is costly but only weakly selected against.\\

\noindent When $A < r$, depth is chosen by candidates if understanding is valued sufficiently ($B > \mathcal{B}_1$) \textit{and} the test is accurate enough. There are two interesting regimes in the intersection space between $B > \mathcal{B}_1$ (the negatively sloped line) and $A < r$ (the vertical line). 
In the first regime, $\mathcal{B}_1 < B < \mathcal{B}_0$, the high-potential applicants learn to go deep, and recruiters learn to hire applicants with many papers who pass the test.
In the second regime, $B > \mathcal{B}_0$, recruiters also learn to hire few-paper applicants who pass, which rewards depth by medium-potential applicants.
In this region, depth by medium-potential applicants dominates at low $A$, where the high-potential applicants go shallow (reverse sorting).
Depth by high-potential applicants dominates as $A$ approaches $r$.
At low test accuracy ($\tau = 0.5$), the high-potential applicants rarely go deep, and the medium-potential applicants do so only when $B$ is very high and $A$ is low.\\

\begin{figure}
    \centering
    \includegraphics[width=0.75\linewidth]{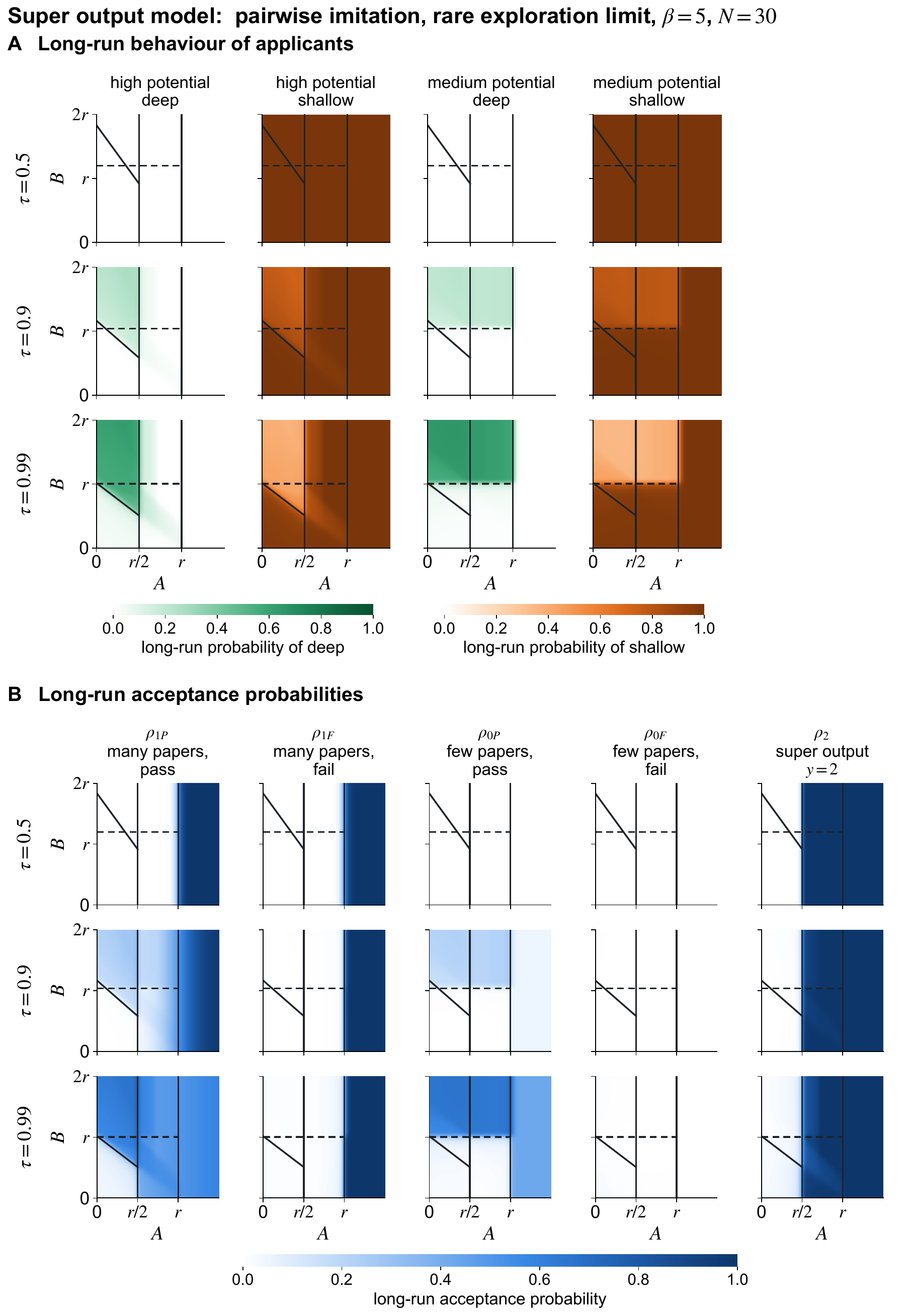}
    \caption{\textbf{Long-run applicant and recruiter behavior in a learning process (extended model).} 
    We study the same process as the one in Figure \ref{fig:baseline-evolution} but for the super-output extension model. Panel \textbf{A} stays the same as Figure \ref{fig:baseline-evolution}. Here, in panel \textbf{B}, we plot the long-run likelihood that recruiters accept after one of five outcomes of an evaluation $(y,t) \in \{(1,\Pass),(1,\Fail),(0,\Pass),(0,\Fail), (2,x)\}$, where $x$ can either be Pass or Fail. In our super-output extension model, we assume that recruiters do not pay attention to the test outcome if they see $y=2$. This is because seeing this outcome automatically implies that the applicant is high-potential and shallow. We plot these likelihoods as we vary the parameters $A$ and $B$ in the horizontal and vertical axes, respectively, and the accuracy of the test, $\tau$, in rows. Game parameters are the same as \textbf{Figure} \ref{fig:equilibrium-regions}. The population size is set at $N=30$ and the strength of imitation, at $\beta = 5$.}
    \label{fig:super-evolution}
\end{figure}

\noindent Next, we discuss the learning outcomes in the extended `super-output' model. \textbf{Figure}~\ref{fig:super-evolution} shows the long-run learned behaviors.
Overall, letting high-potential applicants produce super-output ($y=2$) by going shallow sharply reduces depth.
Recruiters accept $y=2$ almost exactly when it pays ($A \geqslant r/2$), so the threshold beyond which depth disappears moves from $A=r$ to $A=r/2$.
For $\tau=0.5$, neither type goes deep anywhere.
At higher test accuracy, high-potential applicants study deeply for $A<r/2$ and high $B$, but less often than in the baseline.
The cause is a weak point in the learning process.
When high-potential applicants go deep, no one produces $y=2$. The recruiter population can neutrally drift into accepting it at no cost (provided they see it, at all).
Once they do, high-potential applicants do better by going shallow, and as a result, depth unravels even where it is an underlying equilibrium that supports it.
Depth is lost not because it stops paying, but because nothing protects the rule of rejecting super-output at the outset.
Indeed, when we repeat the simulations with recruiters barred from accepting super-output, depth by high-potential applicants rises above even its baseline level (we do not present this result in main text; please find our code in the repository).
Finally, for $r/2 \leqslant A < r$, depth survives only among medium-potential applicants, at high test accuracy and $B>\mathcal{B}_0$, alongside super-output hiring of high-potential applicants (reverse sorting).\\

\section{Discussion}\label{sec:discussion}

Our model links applicants' study choices to recruiters' hiring incentives. The model's predicted outcomes depend on the value recruiters place on output and understanding, the cost of study, and the accuracy of assessment. These factors shape who studies deeply and who gets hired. For departments seeking to encourage mathematical understanding, the model points to two priorities: how they reward understanding and how reliably they assess it. Here, we discuss some important inferences from the analysis we have performed on our model.\\

\noindent \textbf{(i) Valuing understanding does not automatically guarantee adoption of deep study.} One of our findings highlights that valuing understanding does not, by itself, create an incentive to acquire it.
Consider a market in which every strategic applicant studies shallowly.
Recruiters know this, either from reasoning or from experience.
They therefore know that everyone who passes the test got through by luck, so a pass carries no information about understanding.
If output alone does not justify hiring (i.e., $A$ is not high enough), recruiters reject every applicant, no matter how highly they value understanding ($B$).
Now suppose a single applicant studies deeply.
That applicant passes the test, but a pass is all the recruiter sees, and it looks exactly like the lucky passes of shallow applicants.
The recruiter has no way to tell them apart, and so they reject both.
Deep study therefore brings no reward, and no applicant has a reason to change course.
A pass means nothing because no one studies deeply, and no one studies deeply because a pass means nothing.
The trap persists even though recruiters interpret every signal correctly.\\

\noindent \textbf{(ii) The stability of the deep-study equilibria depends on institutional expectations.} Every pure-strategy equilibrium with deep study coexists with an equilibrium in which all strategic applicants study shallowly (\textbf{Figure} \ref{fig:equilibrium-regions}).
Institutional expectations determine which of the two outcomes takes hold.
The same valuation of output, $A$, and understanding, $B$, and the same test accuracy, $\tau$, can support both a market in which no one studies deeply and one in which many do.
What separates them is not these fundamentals but what each side expects of the other.
Recruiters' expectations shape how they read a pass.
If they expect applicants to study deeply, a pass signals understanding and they hire on it.
If they expect shallow study, a pass signals only luck and they reject.
Applicants' expectations shape whether deep study is worth its cost.
If applicants expect recruiters to hire those who pass, deep study pays.
If they expect recruiters to reject everyone, it does not.
Each side's expectation is confirmed by the other's behavior, so both outcomes sustain themselves.
No single applicant or department can move the market alone.
A department that starts rewarding passes at first hires only lucky applicants.
Applicants change course only once enough departments do (a lone applicant who studies deeply also goes unrewarded).
Escaping the trap therefore requires coordination.
Departments must commit together, and in advance, to rewarding demonstrated understanding.
Our learning simulations suggest that such coordination is hard to reach.
In the rare-exploration limit we study, only one population changes at a time, and a department or applicant that changes alone is penalized by the other side's current behavior.
Populations therefore often settle where no one studies deeply, even when deep study could be sustained, especially when tests are less accurate or when AI amplifies high-potential output.
Faster exploration, which allows several departments and applicants to change at once, may make coordination easier.\\

\noindent \textbf{(iii) Existence of a reverse-sorting equilibrium.} One of the more surprising outcomes from our model is the reverse-sorting equilibrium, in which medium-potential applicants study deeply while high-potential applicants rely on AI (equilibrium $\mathbf{E}_4$).
The logic rests on what each signal comes to mean.
High-potential applicants produce many papers whether they study deeply or not.
Medium-potential applicants who study deeply produce few papers but pass the test.
Suppose recruiters expect high-potential applicants to study shallowly.
Then every applicant with many papers is a shallow worker, and recruiters rationally reject them, even those who pass.
A high-potential applicant who studies deeply gains nothing, because they cannot avoid producing many papers and so cannot escape this inference.
Medium-potential applicants face the opposite situation.
Few papers combined with a pass now marks someone who studied deeply, so recruiters hire on it, and deep study pays.
High-potential applicants are locked out by their own productivity.
In this equilibrium, abundant output has become a sign of shallowness rather than ability.\\

\noindent \textbf{(iv) Applicants' perception of test accuracy matters.} In our model, applicants know the accuracy of the test exactly.
In practice, they may only hold a belief about it.
Our conditions show what would happen in that case.
A high-potential applicant studies deeply only if $\tau W \geqslant \delta_H$, and this decision rests on how accurate the applicant believes the test to be.
If applicants underestimate the test's accuracy, they study shallowly even when the test would catch them.
A rigorous test that applicants consider easy to pass therefore deters no applicant.
Depth requires tests that are both accurate and known to be accurate.
This gives transparent assessment standards real value.\\

\noindent \textbf{(v) A more accurate test by itself does not encourage understanding.} Better assessment alone does not protect understanding.
When output by itself justifies hiring, recruiters hire applicants with many papers even when they fail the test.
Depth then disappears regardless of how accurate the test is.
A test shapes behavior only if a failed test actually costs the applicant the job.
Investment in assessment therefore matters only alongside hiring rules that use the assessment's results.\\

\noindent \textbf{(vi) Hiring more does not guarantee hiring better understanding.}
Equilibria differ both in how many applicants are hired and in how many of the hired understand their field.
Hiring every applicant who passes ($\mathbf{E}_3$) raises the hiring rate but also admits low-potential applicants who passed by luck.
Hiring only applicants with many papers who pass ($\mathbf{E}_2$) hires fewer people but screens out low-potential applicants entirely, at the cost of admitting some medium-potential AI users.
Which rule yields more understanding per hire depends only on the applicant pool, not on test accuracy (\textbf{Table}~\ref{tab:hire-understand}).
Hiring every applicant who passes does better when low-potential applicants are rare relative to medium-potential ones: that is, when $\pi_H\pi_L \leqslant \pi_M(\pi_H+\pi_M)$.
When low-potential applicants are common, hiring only many-paper applicants who pass does better.\\

\noindent \textbf{(vii) Ongoing reforms through the lens of the model.}
Some of the ongoing reforms discussed in the introduction of our paper can be conceptually mapped to elements in our model. Oral defenses and chalk talks \citep{summit2026,simons2026} raise test accuracy $\tau$, which helps both sides.
It makes a pass more informative to recruiters and makes deep study more rewarding to applicants.
Raising the value placed on understanding, $B$, helps only the recruiter's side.
Caps on publication counts \citep{simons2026,arxiv2026} lower the weight on output, $A$, which moves departments away from the output-dominant regime where test outcomes are ignored.\\

\noindent \textbf{(viii) Commit in advance to rules for output no one yet produces.}
In the super-output model, depth erodes through neutral drift in the learning process.
When $A<r/2$, super-output alone does not secure a job, so high-potential applicants learn to study deeply.
Recruiters then never see a super-output application, and accepting or rejecting one makes no difference to their payoff.
A recruiter strategy that accepts super-output can therefore spread purely by chance.
Once it does, high-potential applicants switch to shallow study and abandon depth.
When recruiters are committed to rejecting super-output, even while no one produces it, high-potential applicants study deeply more often.
In other words, stating in advance that output volume alone will not secure a job may therefore protect depth more effectively than better tests when super-output is possible.\\

\subsection{Related work}

Our model builds on job-market signaling \citep{spence1973} and career-concerns models, in which employers infer ability from past output \citep{holmstrom1999, gibbons1992, dewatripont1999}. We use evolutionary dynamics to study which outcomes emerge through learning. This builds on past work applying evolutionary game theory to human social systems \citep{johnson2011,hoffmansignal2018}. For the goal of making predictions about a job-market-signaling model, the present paper applies evolutionary dynamics (a computational method) in addition to static game theory (an analytical method), while past work has traditionally relied solely on the latter. The application of evolutionary dynamics to job-market signaling is relatively unexplored, although not entirely new \citep{hofbauer2025}.\\

\noindent Recent studies examine how AI changes the information conveyed by output. Galdin and Silbert \citep{galdin2025} estimate a structural model using data from a digital-labor platform. In their counterfactual, employers become less able to identify high-ability workers when AI makes written applications uninformative. Wang \citep{wang2026} models the pre-doctoral labor market. The model predicts that AI can raise routine research output while weakening its value as a signal of talent. Cetemen and Ozdenoren \citep{cetemen2026} study a setting with abundant AI and opportunities to sell output on the side. They find that tradable AI output reverses the classic signaling rat race:
effort no longer signals ability, and every worker type under-invests in skill. Our model builds on this body of work by adding understanding as a second applicant-side variable on top of output. \\

\noindent AI's effects can also vary across settings. Using theory and empirical evidence, Cowgill et al. \citep{cowgill2026} find that AI use by job-market and startup-investing applicants can both increase and decrease screening accuracy depending on the circumstances. Yao \citep{yao2026b} analyzes competitive-programming data, finding that AI-style practice is associated with weaker skill growth in some groups and stronger skill performance in others. In contrast, our model studies the idealized setting where AI use by applicants always results in a missed opportunity to build understanding.

\subsection{Limitations}

The model simplifies both learning and output.
\emph{(i)} Understanding is binary: deep study always produces it, and shallow study never does.
\emph{(ii)} Output is also deterministic: different applicant types can produce exactly the same output. This creates an idealized form of ambiguity in the hiring signal, particularly in the baseline model. The super-output extension relaxes this in part. There, a high-potential applicant produces more output than a medium-potential applicant whenever both choose the same trajectory. Yet a deep-studying high-potential applicant and a shallow-studying medium-potential applicant still produce the same output. 
\emph{(iii)} In this model, study costs, applicant composition, and test accuracy are fixed externally and cannot respond to changes in behavior or hiring practices. Future models could allow uncertain learning, variable output, and feedback between these factors. \textit{(iv)} Finally, our model does not allow applicants to gain understanding while using AI for research. One could potentially study a model where interactions with AI when writing a paper lead to gaining a deep understanding of the field. We have excluded this case from our model.

\noindent \textbf{AI disclosure statement.} We have used Claude Opus 5.5 for helping with writing computer code and checking the equilibrium analysis of our model. We have also used GPT-6 Astra and Claude Fable 5.1 for helping us review and revise our work.\\

\noindent \textbf{Code availability.} The code corresponding to this manuscript can be found in the online repository: \href{https://github.com/Saptarshi07/Effort-Hiring-AI-Game}{https://github.com/Saptarshi07/Effort-Hiring-AI-Game}.\\

\noindent \textbf{Acknowledgments.} We are grateful to Harvard University for providing us with AI access.

\clearpage
\printbibliography[title={References}]

\clearpage
\appendix
\section{Methods}\label{sec:methods}
\subsection*{Imitation and rare exploration in applicant and recruiter populations}

\paragraph{Populations and payoffs.}
We model learning in three finite populations, each of size $N$: high-potential applicants ($H$), medium-potential applicants ($M$), and recruiters ($R$).
Each applicant chooses an action $a \in \{D, S\}$ (deep or shallow).
Each recruiter uses a pure acceptance rule $\rho$, which specifies whether to accept ($\rho_o = 1$) or reject ($\rho_o = 0$) after each observation $o$.
In the baseline model, the observations are $(1,P)$, $(1,F)$, $(0,P)$, and $(0,F)$, giving $2^4 = 16$ rules.
In the super-output model, recruiters additionally respond to $y = 2$ through $\rho_2$, giving $2^5 = 32$ rules.
We let recruiters condition on $y = 2$ alone, because super-output is produced only by high-potential applicants who went shallow; the test result therefore carries no further information about understanding.\\

\noindent Low-potential applicants are non-strategic and are not a learning population.
They are a fixed share $\pi_L$ of every applicant pool and produce $(y,u) = (0,0)$, and pass the test with probability $1-\tau$. Their effect enters the dynamics only through the recruiters' payoffs. A recruiter who accepts few-paper applicants also hires low-potential ones, which lowers its payoff by $ r\,\pi_L\bigl((1-\tau)\,\rho_{0P} + \tau\,\rho_{0F}\bigr)$.\\

\noindent Payoffs are expected values against the current composition of the \emph{other} populations.
An applicant of type $\theta \in \{H, M\}$ playing $a$ earns
\[
U_\theta(a) = W \Pr(\text{hired} \mid \theta, a) - \delta_\theta \mathbf{1}\{a = D\},
\]
where the probability of being hired is averaged over the test outcome and the recruiters' rule.
A recruiter using rule $\rho$ against applicant actions $(a_H, a_M)$ earns
\[
U_R(\rho) = \sum_{\theta \in \{H,M,L\}} \pi_\theta \, \mathbb{E}\bigl[(Ay + Bu - r)\,\rho_{o}\bigr],
\]
where the expectation is over the observation $o$ produced by type $\theta$.
Population sizes are therefore equal, but recruiters meet the three types in proportions $\pi_H$, $\pi_M$, $\pi_L$, which are exogenously set for the simulation.
No agent's payoff depends on the composition of its own population.

\paragraph{Imitation and exploration.}
In each time step, one population is chosen with probability $1/3$.
With probability $1-\varepsilon$, a randomly chosen focal agent compares itself with a randomly chosen role model from the same population.
It adopts the role model's strategy with probability $\bigl[1 + e^{-\beta (U_{\text{model}} - U_{\text{focal}})}\bigr]^{-1}$, where $\beta \geqslant 0$ is the selection strength.
This is the pairwise-comparison (Fermi) rule \cite{traulsen2006stochastic}.
With probability $\varepsilon$, the focal agent instead explores: it switches to a strategy drawn uniformly from the other strategies available to its population.

\paragraph{Rare exploration.}
We analyze the limit $\varepsilon \to 0$ \cite{fudenberg2006imitation}.
Exploration is then so rare that each new strategy either takes over its population or disappears before the next one arises.
The process therefore spends almost all its time in monomorphic states $s = (a_H, a_M, \rho)$, of which there are $4 \times 16 = 64$ in the baseline model and $4 \times 32 = 128$ in the super-output model.
Consider a single explorer whose payoff exceeds that of the residents by $\Delta$.
Because its own population's composition does not affect payoffs, $\Delta$ stays constant while the new strategy spreads.
Its probability of taking over the population is exactly
\[
\phi(\Delta) =
\begin{cases}
\dfrac{1 - e^{-\beta\Delta}}{1 - e^{-N\beta\Delta}}, & \beta\Delta \neq 0,\\[8pt]
1/N, & \beta\Delta = 0.
\end{cases}
\]
From state $s$, the process moves to a state that differs only in $a_H$ with probability $\tfrac13 \phi(\Delta_H)$, and similarly for $a_M$.
It moves to a state that differs only in the recruiter rule, $\rho \to \rho'$, with probability $\tfrac13 \phi(\Delta_R) / (|\mathcal{R}| - 1)$, where $|\mathcal{R}|$ is the number of recruiter rules.
Here, each $\Delta$ is the payoff of the new strategy minus that of the resident, evaluated at $s$.
We compute the unique stationary distribution $\Pi$ of this chain exactly by solving a linear system.

\paragraph{Reported quantities.}
We report two kinds of quantities, with no classification of states into equilibria.
The first is the long-run probability that each applicant population plays each action: for example, $\Pr(a_H = D) = \sum_{s:\, a_H = D} \Pi(s)$.
The second is the long-run probability that recruiters accept after each observation, $\sum_s \Pi(s)\, \rho_o(s)$, where $\rho_o(s)$ is the response to $o$ under the rule in state $s$.
Two rules that differ only after an observation that no current applicant produces earn identical payoffs.
Switching between them is neutral ($\phi = 1/N$), which is the learning counterpart of the freedom in off-path beliefs.
Observations that occur only rarely, such as a passed test by a low-potential applicant, are not neutral; they are only weakly selected.

\paragraph{Parameters for learning simulations.}
Unless stated otherwise, $N = 30$, $\beta = 5$, $\pi_H = 0.3$, $\pi_M = 0.5$, $\pi_L = 0.2$, $\delta_H = 0.2$, $\delta_M = 0.4$, and $r = W = 1$.
We use $\tau \in \{0.5, 0.9, 0.99\}$ and a $101 \times 101$ grid with $A \in [0, 1.5]$ and $B \in [0, 2]$.\\

\end{document}